\documentclass[11pt,a4paper]{article}
\usepackage[hyperref]{naaclhlt2019}
\usepackage{times}
\usepackage{latexsym}
\usepackage{booktabs}
\usepackage{hhline}
\usepackage{amsmath}
\usepackage{graphicx, subcaption}
\usepackage{enumitem}
\usepackage{multirow}
\usepackage{ctable}
\usepackage{array}
\usepackage{url}

\usepackage{etoolbox}
\makeatletter
\patchcmd\@combinedblfloats{\box\@outputbox}{\unvbox\@outputbox}{}{\errmessage{\noexpand patch failed}}
\makeatother

\aclfinalcopy % Uncomment this line for the final submission
\renewcommand\vec[1]{\overrightarrow{#1}}
\newcommand\cev[1]{\overleftarrow{#1}}

\title{Modeling Acoustic-Prosodic Cues\\ for Word Importance Prediction in Spoken Dialogues}

\author{Sushant Kafle, Cecilia O. Alm, Matt Huenerfauth \\
    Rochester Institute of Technology, Rochester NY \\ {\tt \{sxk5664,matt.huenerfauth,coagla\}@rit.edu}}
\date{}

\begin{document}
\maketitle
\begin{abstract}
 Prosodic cues in conversational speech aid listeners in discerning a message. We investigate whether acoustic cues in spoken dialogue can be used to identify the importance of individual words to the meaning of a conversation turn. Individuals who are Deaf and Hard of Hearing often rely on real-time captions in live meetings. Word error rate, a traditional metric for evaluating automatic speech recognition (ASR), fails to capture that some words are more important for a system to transcribe correctly than others. We present and evaluate neural architectures that use acoustic features for 3-class word importance prediction. Our model performs competitively against state-of-the-art text-based word-importance prediction models, and it demonstrates particular benefits when operating on imperfect ASR output. 
 
\end{abstract}

\section{Introduction}
\label{sec:introduction}
 Not all words are equally important to the meaning of a spoken message. Identifying the importance of words is useful for a variety of tasks including text classification and summarization \cite{hong2014improving,yih2007multi}. Considering the relative importance of words can also be valuable when evaluating the quality of output of an automatic speech recognition (ASR) system for specific tasks, such as caption generation for Deaf and Hard of Hearing (DHH) participants in spoken meetings \cite{kafle2016}.
 
 As described by \citeauthor{berke2018} \shortcite{berke2018}, interlocutors may submit audio of individual utterances through a mobile device to a remote ASR system, with the text output appearing on an app for DHH users. With ASR being applied to new tasks such as this, it is increasingly important to evaluate ASR output effectively. Traditional Word Error Rate (WER)-based evaluation assumes that all word transcription errors equally impact the quality of the ASR output for a user. However, this is less helpful for various applications \cite{mccowan2004use,morris2004and}. In particular, \citeauthor{kafle2016} \shortcite{kafle2016} found that metrics with differential weighting of errors based on word importance correlate better with human judgment than WER does for the automatic captioning task. However, prior models based on text features for word importance identification \cite{kafle2017, sheikh2016learning} face challenges when applied to conversational speech:
 
 \begin{itemize}
 \item\textbf{Difference from Formal Texts}: Unlike formal texts, conversational transcripts may lack capitalization or punctuation, use informal grammatical structures, or contain disfluencies (e.g. incomplete words or edits, hesitations, repetitions), filler words, or more frequent out-of-vocabulary (and invented) words \cite{mckeown2005text}.
 
  \item\textbf{Availability and Reliability}: Text transcripts of spoken conversations require a human transcriptionist or an ASR system, but ASR transcription is not always reliable or even feasible, especially for noisy environments, nonstandard language use, or low-resource languages, etc.
 \end{itemize}

\begin{figure}[!t]
\centering
\includegraphics[width=0.48\textwidth]{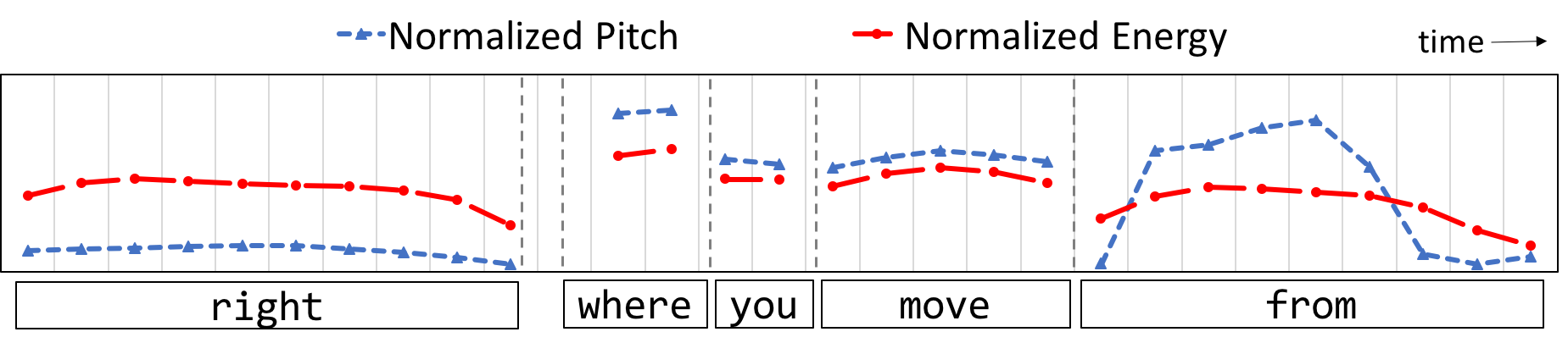}
\caption{Example of conversational transcribed text, \textit{right where you move from}, that is difficult to disambiguate without prosody. The intended sentence structure was: \textit{Right! Where you move from?}}
\label{fig:speechinfo}
\end{figure}

While spoken messages include prosodic cues that focus a listener's attention on the most important parts of the message \cite{frazier2006prosodic}, such information may be omitted from a text transcript, as in Figure \ref{fig:speechinfo}, in which the speaker pauses after ``right" (suggesting a boundary) and uses rising intonation on ``from" (suggesting a question). Moreover, there are application scenarios where transcripts of spoken messages are not always available or fully reliable. In such cases, models based on a speech signal (without a text transcript) might be preferred. 

With this motivation, we investigate modeling acoustic-prosodic cues for predicting the importance of words to the meaning of a spoken dialogue. Our goal is to explore the versatility of speech-based (text-independent) features for word importance modeling. In this work, we frame the task of word importance prediction as sequence labeling and utilize a bi-directional Long Short-Term Memory (LSTM)-based neural architecture for context modeling on speech.

\section{Related Work}
Many researchers have considered how to identify the importance of a word and have proposed methods for this task. Popular methods include frequency-based unsupervised measures of importance, such as Term Frequency-Inverse Document Frequency (TF-IDF), and word co-occurrence measures \cite{hacohen2005automatic,matsuo2004keyword}, which are primarily used for extracting relevant keywords from text documents. Other supervised measures of word importance have been proposed \cite{liu2011,liu2004text,hulth2003improved,sheeba2012improved, kafle2017} for various applications. Closest to our current work, researchers in \cite{kafle2017}  described a neural network-based model for capturing the importance of a word at the sentence level. Their setup differed from traditional importance estimation strategies for document-level keyword-extraction, which had treated each \textit{word} as a \textit{term} in a document such that all \textit{words} identified by a \textit{term} received a uniform importance score, without regard to context. Similar to our application use-case, the model proposed by \citeauthor{kafle2017} \shortcite{kafle2017} identified word importance at a more granular level, i.e. sentence- or utterance-level. However, their model operated on human-generated transcripts of text. Since we focus on real-time captioning applications, we prefer a model that can operate without such human-produced transcripts, as discussed in Section \ref{sec:introduction}.

Previous researchers have modeled prosodic cues in speech for various applications \cite{tran2017joint, brenier2005detection, xie2009integrating}.  For instance, in automatic prominence detection, researchers predict regions of speech with relatively more spoken stress \cite{wang2007acoustic, brenier2005detection, tamburini2003prosodic}. Identification of prominence aids automatically identifying content words \cite{wang2007acoustic}, a crucial sub-task of spoken language understanding \cite{beckman2000tagging, mishra2012word}. Moreover, researchers have investigated modeling prosodic patterns in spoken messages to identify syntactic relationships among words \cite{price1991use, tran2017joint}. In particular, \citeauthor{tran2017joint} demonstrated the effectiveness of speech-based features in improving the constituent parsing of conversational speech texts. In other work, researchers investigated prosodic events to identify important segments in speech, useful for producing a generic summary of the recordings of meetings \cite{xie2009integrating, murray2005extractive}. At the same time, prosodic cues are also challenging in that they serve a range of linguistic functions and convey affect. We investigate models applied to spoken messages at a dialogue-turn level, for predicting the importance of words for understanding an utterance.

\section{Word Importance Prediction}
\label{sec:model-architecture}
For the task of word importance prediction, we formulate a sequence labeling architecture that takes as input a spoken dialogue turn utterance with word-level timestamps\footnote{For the purposes of accurately evaluating efficacy of speech-based feature for word importance, we currently make use of high-quality human-annotated word-level timestamp information in our train/evaluation corpus; in the future, speech tokenization could  be automated.}, and assigns an importance label to every spoken word in the turn using a bi-directional LSTM architecture \cite{huang2015bidirectional, lample2016neural}.

\begin{figure*}[t!]
\centering
\includegraphics[width=0.8\textwidth]{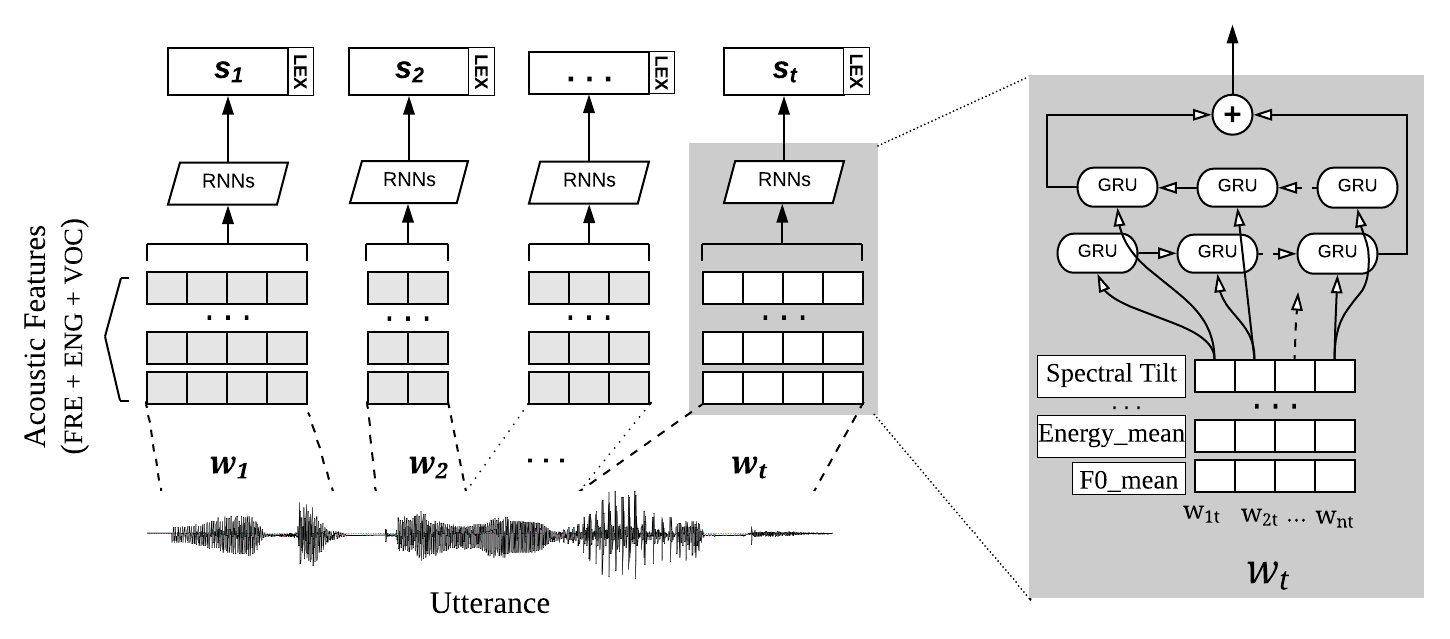}
\caption{Architecture for feature representation of spoken words using time series speech data. For each spoken word ($w$) identified by a word-level timestamp, a fixed-length interval window ($\tau$) slides through to get $n = time(w)/\tau$ sub-word interval segments. Using an RNN network, a word-level feature ($s$), represented by a fixed-length vector, is extracted using the features from a variable-length sub-word sequence.}
\label{fig-feat-represent}
\vspace{-4mm}
\end{figure*}

\begin{align} \label{eq:bi_lstm_1}
    \vec{h_t} = LSTM(s_t, \vec{h_{t-1}}) \\
    \label{eq:bi_lstm_2}
    \cev{h_t} = LSTM(s_t, \cev{h_{t-1}})
\end{align}

The word-level timestamp information is used to generate an acoustic-prosodic representation for each word ($s_t$) from the speech signal. Two LSTM units, moving in opposite directions through these word units ($s_t$) in an utterance, are then used for constructing a context-aware representation for every word. Each LSTM unit takes as input the representation of the word ($s_t$), along with the hidden state from the previous time step, and each outputs a new hidden state. At each time step, the hidden representations from both LSTMs are concatenated $h_t=[\vec{h_t};\cev{h_t}]$, in order to obtain a contextualized representation for each word. This representation is next passed through a projection layer (details below) to the final prediction for a word.

\subsection{Importance as Ordinal Classification}
\label{sec:projection_layers}

We define word importance prediction as the task of classifying the words into one of the many importance classes, e.g., high importance (\textsc{hi}), medium importance (\textsc{mid}) and low importance (\textsc{low}) (details on Section \ref{sec:exp-setup}). These importance class labels have a natural \textit{ordering} such that the cost of misclassification is not uniform e.g., incorrect classification of \textsc{hi} class for \textsc{li} class (or vice-versa) will have higher error cost than classification of \textsc{hi} class for \textsc{mi}. Considering this ordinal nature of the importance class labels, we investigate  different projection layers for output prediction: a softmax layer for making local importance prediction (\textsc{softmax}), a relaxed softmax tailored for ordinal classification (\textsc{ord}), and a linear-chain conditional random field (\textsc{crf}) for making a conditioned decision on the whole sequence.\\

\noindent \textbf{Softmax Layer}. For the \textsc{softmax}-layer, the model predicts a normalized distribution over all possible labels ($L$) for every word conditioned on the hidden vector ($h_t$).
\vspace{2mm}

\noindent \textbf{Relaxed Softmax Layer}. In contrast, the \textsc{ord}-layer uses a standard sigmoid projection for every output label candidate, without subjecting it to normalization. The intuition is that rather than learning to predict one label per word, the model predicts multiple labels. For a word with label $l \in L$, all other labels ordinally less than $l$ are also predicted. Both the softmax and the relaxed-softmax models are trained to minimize the categorical cross-entropy, which is equivalent to minimizing the negative log-probability of the correct labels. However, they differ in how they make the final prediction: Unlike the \textsc{softmax} layer which considers the most probable label for prediction, the \textsc{ord}-layer uses a special ``scanning" strategy \cite{ordinal} -- where for each word, the candidate labels are scanned from low to high (ordinal rank), until the score from a label is smaller than a threshold (usually 0.5) or no labels remain. The last scanned label with score greater than the threshold is selected as the output.
\vspace{2mm}

\noindent \textbf{CRF Layer}. The \textsc{crf}-layer explores the possible dependence between the subsequent importance label of words. With this architecture, the network looks for the most optimal path through all possible label sequences to make the prediction. The model is then optimized by maximizing the score of the correct sequence of labels, while minimizing the possibility of all other possible sequences.

\noindent Considering each of these different projection layers, we investigate different models for the word importance prediction task. Section \ref{sec:acoustic_feature_representation} describes our architecture for acoustic-prosodic feature representation at the word level, and Sections \ref{sec:exp-setup} and \ref{sec:results} describe our experimental setup and subsequent evaluations.

\section{Acoustic-Prosodic Feature Representation}
\label{sec:acoustic_feature_representation}

Similar to familiar feature-vector representations of words in a text e.g., word2vec \cite{word2vec} or GloVe \cite{glove}, various researchers have investigated vector representations of words based on speech. In addition to capturing acoustic-phonetic properties of speech \cite{he2016multi,chung2016audio}, some recent work on acoustic embeddings has investigated encoding semantic properties of a word directly from speech \cite{speech2vec}. In a similar way, our work investigates a speech-based feature representation strategy that considers prosodic features of speech at a sub-word level, to learn a word-level representation for the task of importance prediction in spoken dialogue.

\subsection{Sub-word Feature Extraction}
\label{acoustic-feats}
We examined four categories of features that have been previously considered in computational models of prosody, including: pitch-related features (10), energy features (11), voicing features (3) and spoken-lexical features (6):
\vspace{2mm}

\noindent $\bullet$ \textbf{Pitch (\textsc{freq}) and Energy (\textsc{eng}) Features:} Pitch and energy features have been found effective for modeling intonation and detecting emphasized regions of speech \cite{brenier2005detection}. From the pitch and energy contours of the speech, we extracted: minimum, time of minimum, maximum, time of maximum, mean, median, range, slope, standard deviation and skewness. We also extracted RMS energy from a mid-range frequency band (500-2000 Hz), which has been shown to be useful for detecting prominence of syllables in speech \cite{tamburini2003prosodic}.
\vspace{2mm}

\noindent $\bullet$ \textbf{Spoken-lexical Features (\textsc{lex}):} We examined spoken-lexical features, including word-level spoken language features such as duration of the spoken word, the position of the word in the utterance, and duration of silence before the word. We also estimated the number of syllables spoken in a word, using the methodology of \citeauthor{de2009praat} \shortcite{de2009praat}. Further, we considered the per-word average syllable duration and the per-word articulation rate of the speaker (number of syllables per second).
\vspace{2mm}

\noindent $\bullet$ \textbf{Voicing Features (\textsc{voc}):} As a measure of voice quality, we investigated spectral-tilt, which is represented as (H1 - H2), i.e. the difference between the amplitudes of the first harmonic (H1) and the second harmonic (H2) in the Fourier Spectrum. The spectral-tilt measure has been shown to be effective in characterizing glottal constriction \cite{voicequal06}, which is important in distinguishing voicing characteristics, e.g. whisper \cite{itoh2001acoustic}.  We also exmined other voicing measures, e.g. Harmonics-to-Noise Ratio and Voiced Unvoiced Ratio. 
\vspace{2mm}

\noindent In total, we extracted 30 features using Praat \cite{praat}, as listed above. Further, we included speaker-normalized (\textsc{znorm}) version of the features. Thereby, we had a total of 60 speech-based features extracted from sub-word units.

\subsection{Sub-word to Word-level Representation}
The acoustic features listed above were extracted from a 50-ms sliding window over each word region with a 10-ms overlap. In our model, each word was represented as a sequence of these sub-word features with varying lengths, as shown in Figure \ref{fig-feat-represent}. To get a feature representation for a word, we utilized a bi-directional Recurrent Neural Network (RNN) layer on top of the sub-word features. The spoken-lexical features were then concatenated to this word-level feature representation to get our final feature vectors. For this task, we utilized Gated Recurrent Units (GRUs) \cite{gru} as our RNN cell, rather than LSTM units, due to better performance observed during our initial analysis.

\section{Experimental Setup}
\label{sec:exp-setup}

We utilized a portion of the Switchboard corpus \cite{switchboard} that had been manually annotated with word importance scores, as a part of the Word Importance Annotation project \cite{kafle2017}. That annotation covers 25,048 utterances spoken by 44 different English speakers, containing word-level timestamp information along with a numeric score (in the range of [0, 1]) assigned to each word from the speakers. These numeric importance scores have three natural ordinal ranges {[0 - 0.3), [0.3, 0.6), [0.6, 1]} that the annotators had used during the annotation to indicate the importance of a word in understanding an utterance. The ordinal range represents low importance (\textsc{li}), medium importance (\textsc{mi}) and high importance (\textsc{hi}) of words, respectively.

Our models were trained and evaluated using this data, treating the problem as a ordinal classification problem with the labels ordered as (\textsc{li} $<$ \textsc{mi} $<$ \textsc{hi}). We created a 80\%, 10\% and 10\% split of our data for training, validation, and testing. The prediction performance of our model was primarily evaluated using the Root Mean Square (RMS) measure, to account for the ordinal nature of labels. Additionally, our evaluation includes F-score and accuracy results to measure classification performance. As our baseline, we used various text-based importance prediction models trained and evaluated on the same data split, as described in Section \ref{sec:comparisontext}.

For training, we explored various architectural parameters to find the best-working setup for our models: Our input layer of GRU-cells, used as word-based speech representation, had a dimension of 64. The LSTM units, used for generating contextualized representation of a spoken word, had a dimension of 128. We used the Adam optimizer with an initialized learning rate of $0.001$ for training. Each training batch had a maximum of 20 dialogue-turn utterances, and the model was trained until no improvement was observed in 7 consecutive iterations.

\section{Experiments}
\label{sec:results}
Tables \ref{tbl:projection_eval}, \ref{tbl:speech_ablation} and \ref{tbl:asr_eval} summarize the performance of our models on the word importance prediction task. The performance scores reported in the tables are the average performance across 5 different trials, to account for possible bias due to random initialization of the model.

\subsection{Comparison of the Projection Layers}
We compared the efficacy of the learning architecture's three projection layers (Section \ref{sec:projection_layers}) by training them separately and comparing their performance on the test corpus. Table \ref{tbl:projection_eval} summarizes the results of this evaluation. \\

\begin{table}[ht!]
    \centering
    \begin{tabular}{llll}
    \hline
    \multicolumn{1}{l|}{Model} & \multicolumn{1}{l|}{ACC} & \multicolumn{1}{l|}{F1} & \multicolumn{1}{l}{RMS} \\ \hhline{====}
    \multicolumn{1}{l|}{\textsc{lstm-crf}}     & \multicolumn{1}{l|}{64.22}       &  \multicolumn{1}{l|}{56.31} & \multicolumn{1}{l}{75.21}        \\ \hline
    \multicolumn{1}{l|}{\textsc{lstm-softmax}} & \multicolumn{1}{l|}{ \textbf{65.66} } & \multicolumn{1}{l|}{ 57.34 } & \multicolumn{1}{l}{ 74.08 }        \\ \hline
    \multicolumn{1}{l|}{\textbf{\textsc{lstm-ord}}} & \multicolumn{1}{l|}{63.72}       & \multicolumn{1}{l|}{ \textbf{57.58} } & \multicolumn{1}{l}{ \textbf{68.21} }        \\ \hline
    
    \end{tabular}
    \caption{Performance of our speech-based models on the test data under different projection layers. Best performing scores highlighted in \textbf{bold}.}
    \label{tbl:projection_eval}
    \vspace{-3mm}
\end{table}

\noindent \textbf{Results and Analysis}: The \textsc{lstm-softmax}-based and \textsc{lstm-crf}-based projection layers had nearly identical performance; however, in comparison, the \textsc{lstm-ord} model had better performance with significantly lower RMS score than the other two models. This suggests the utility of the ordinal constraint present in the \textsc{ord}-based model for word importance classification.

\subsection{Ablation Study on Speech Features}
To compare the effect of different categories of speech features on the performance of our model, we evaluated variations of the model by removing one feature group at a time from the model during training.  Table \ref{tbl:speech_ablation} summarizes the results of the experiment.\\

\begin{table}[!ht]
\centering
\begin{tabular}{llll}
\hline
\multicolumn{1}{l|}{Model} & \multicolumn{1}{l|}{ACC} & \multicolumn{1}{l|}{F1} & \multicolumn{1}{l}{RMS} \\ \hhline{====}
\multicolumn{1}{l|}{\textit{speech-based}} & \multicolumn{1}{l|}{63.72} & \multicolumn{1}{l|}{57.58} & \multicolumn{1}{l}{68.21} \\ \hline
\multicolumn{1}{l|}{\hspace{5mm} \textbf{-- \textsc{eng}}} & \multicolumn{1}{l|}{62.24$^\dagger$} & \multicolumn{1}{l|}{55.67$^\dagger$} & \multicolumn{1}{l}{71.14} \\ \hline
\multicolumn{1}{l|}{\hspace{5mm} \textbf{-- \textsc{freq}}} & \multicolumn{1}{l|}{63.25} & \multicolumn{1}{l|}{57.30} & \multicolumn{1}{l}{69.0} \\ \hline
\multicolumn{1}{l|}{\hspace{5mm} \textbf{-- \textsc{voc}}} & \multicolumn{1}{l|}{62.90} & \multicolumn{1}{l|}{56.84} & \multicolumn{1}{l}{70.5} \\ \hline
\multicolumn{1}{l|}{\hspace{5mm} \textbf{-- \textsc{lex}}} & \multicolumn{1}{l|}{63.37} & \multicolumn{1}{l|}{57.34} & \multicolumn{1}{l}{71.49$^\dagger$} \\ \hline
\multicolumn{1}{l|}{\hspace{5mm} \textbf{-- \textsc{znorm}}} & \multicolumn{1}{l|}{62.04$^\star$} & \multicolumn{1}{l|}{53.86$^\star$} & \multicolumn{1}{l}{72.0$^\star$} \\ \hline
\end{tabular}
\caption{Speech feature ablation study. The minus sign indicates the feature group removed from the model during training. Markers ($\star$ and $\dagger$) indicate the biggest and the second-biggest change in model performance for each metric, respectively.}
\label{tbl:speech_ablation}
\vspace{-3mm}
\end{table}

\noindent \textbf{Results and Analysis:} Omitting speaker-based normalization (\textsc{znorm}) features and omitting spoken-lexical features (\textsc{lex}) resulted in the greatest increase in the overall RMS error (+5.5\% and +4.8\% relative increase in RMS respectively) -- suggesting the discriminative importance of these features for word importance prediction. Further, our results indicated the importance of energy-based (\textsc{eng}) features, which resulted in a substantial drop (-2.4\% relative decrease) in accuracy of the model.

\subsection{Comparison with the Text-based Models}
\label{sec:comparisontext}
In this analysis, we compare our best-performing speech-based model with a state-of-the-art word-prediction model based on text features; this prior text-based model did not utilize any acoustic or prosodic information about the speech signal.  The baseline text-based word importance prediction model used in our analysis is described in \citeauthor{kafle2017} \shortcite{kafle2017}, and it uses pre-trained word embeddings and bi-direction LSTM units, with a CRF layer on top, to make a prediction for each word. 

As discussed in Section \ref{sec:introduction}, human transcriptions are difficult to obtain in some applications, e.g. real-time conversational settings. Realistically, text-based models need to rely on ASR systems for transcription, which will contain some errors. Thus, we compare our speech-based model and this prior text-based model on two different types of transcripts: manually generated or ASR generated. We processed the original speech recording for each segment of the corpus with an ASR system to produce an automatic transcription. To simulate different word error rate (WER) levels in the transcript, we also artificially injected the original speech recording with white-noise and then processed it again with our ASR system. Specifically, we utilized Google Cloud Speech\footnote{\url{https://cloud.google.com/Speech\_API}} ASR with WER$\approx25\%$ on our test data (without the addition of noise) and WER$\approx30\%$ after noise was inserted.  Given our interest in generating automatic captions for DHH users in a live meeting on a turn-by-turn basis (Section 1), we provided the ASR system with the recording for each dialogue-turn individually, which may partially explain these somewhat high WER scores.

The automatically generated transcripts were then aligned with the reference transcript to compare the importance scores. Insertion errors automatically received a label of low importance (\textsc{li}). The WER for each ASR system was computed by performing a word-to-word comparison, without any pre-processing (e.g., removal of filler words).  

\begin{table}[!ht]
\centering
\begin{tabular}{llll}
    \hline
    \multicolumn{1}{l|}{Model} & \multicolumn{1}{l|}{ACC} & \multicolumn{1}{l|}{F1} & \multicolumn{1}{l}{RMS} \\ \hhline{====}
    \multicolumn{1}{l|}{\textit{speech-based}} & \multicolumn{1}{l|}{ 63.72 }       & \multicolumn{1}{l|}{ 57.58 }  & \multicolumn{1}{l}{ 68.21}       \\ \hline
    \multicolumn{1}{l|}{\textit{text-based}}     & \multicolumn{1}{l|}{ 77.81 }       & \multicolumn{1}{l|}{ 73.6 } & \multicolumn{1}{l}{ 54.0 }        \\ \hline
    \multicolumn{1}{l|}{\hspace{5mm} \textbf{\textsc{+ wer}: 0.25}}     & \multicolumn{1}{l|}{ 72.30 }       & \multicolumn{1}{l|}{ 69.04 }   & \multicolumn{1}{l}{ 65.15 }     \\ \hline
    \multicolumn{1}{l|}{\hspace{5mm} \textbf{\textsc{+ wer}: 0.30}}     & \multicolumn{1}{l|}{ 71.84 }       & \multicolumn{1}{l|}{ 67.71 } & \multicolumn{1}{l}{ 68.55 }        \\ \hline
    \end{tabular} 
    \caption{Comparison of our speech-based model with a prior text-based model, under different word error rate conditions.}
    \label{tbl:asr_eval}
    \vspace{-3mm}
\end{table}

\vspace{0.2cm}
\noindent \textbf{Result and Analysis:} Given the significant lexical information available for the text-based model, it would be natural to expect that it would achieve higher scores than would a model based only on acoustic-prosodic features.  As expected, Table \ref{tbl:asr_eval} reveals that when operating on perfect human-generated transcripts (with zero recognition errors), the text-based model outperformed our speech-based model.  However, when operating on ASR transcripts (including recognition errors), the speech-based models were competitive in performance with the text-based models. In particular, prior work has found that WER of $\approx30\%$ is typical for modern ASR in many real-world settings or without good-quality microphones \cite{lasecki2012real, barker2017chime}.  When operating on such ASR output, the RMS error of the speech-based model and the text-based model were comparable.

\section{Conclusion}
\label{conclusion}
Motivated by recent work on evaluating the accuracy of automatic speech recognition systems for real-time captioning for Deaf and Hard of Hearing (DHH) users \cite{kafle2017}, we investigated how to predict the importance of a word to the overall meaning of a spoken conversation turn. In contrast to prior work, which had depended on text-based features, we have proposed a neural architecture for modeling prosodic cues in spoken messages, for predicting word importance. Our text-independent speech model had an F-score of $56$ in a 3-class word importance classification task. Although a text-based model utilizing pre-trained word representation had better performance, acquisition of accurate speech conversation text-transcripts is impractical for some applications. When utilizing popular ASR systems to automatically generate speech transcripts as input for text-based models, we found that model performance decreased significantly. Given this potential we observed for acoustic-prosodic features to predict word importance continued work involves combining both text- and speech-based features for the task of word importance prediction.

\section{Acknowledgements}
This material was based on work supported by the Department of Health and Human Services under Award No. 90DPCP0002-01-00, by a Google Faculty Research Award, and by the National Technical Institute of the Deaf (NTID).

% include your own bib file like this:
%\bibliographystyle{acl}
%\bibliography{naaclhlt2018}
\bibliography{naaclhlt2019}
\bibliographystyle{acl_natbib}

\end{document}